%% file: main.tex
\documentclass[conference]{IEEEtran}
\IEEEoverridecommandlockouts
\usepackage{cite}
\usepackage{amsmath,amssymb,amsfonts}
\usepackage{algorithmic}
\usepackage{graphicx}
\usepackage{textcomp}
\usepackage{xcolor}
\def\BibTeX{{\rm B\kern-.05em{\sc i\kern-.025em b}\kern-.08em
    T\kern-.1667em\lower.7ex\hbox{E}\kern-.125emX}}

\usepackage{hyperref}
\usepackage{acronym}
\usepackage{outlines}
\usepackage{pdfpages}
\usepackage{enumitem}
\usepackage{scrextend}
\usepackage{dblfloatfix}
\usepackage{booktabs}
\usepackage{float}
\usepackage{multirow}

\begin{document}

\title{SoK: Assessing the State of Applied \\ Federated Machine Learning
}

\author{
Tobias M\"uller$^1$, Maximilian St\"abler$^2$, Hugo Gasc\'on$^3$, Frank K\"oster$^2$ and Florian Matthes$^1$\\
\textit{$^1$Technical University of Munich}\\
\textit{$^2$German Aerospace Center (DLR)}\\
\textit{$^3$German Edge Cloud (GEC)}\\
\textit{\{tobias.mueller1, matthes\}@tum.de, \{maximilian.staebler, frank.koester\}@dlr.de, hugo.gascon@gec.io}
}

\maketitle
\input{abstract}
\input{section/introduction.tex}
\input{section/preliminaries}
\input{section/research_process}
\input{section/overview_results.tex}
\input{section/conclusion.tex}

\input{bibliography}
\newpage
\appendix
\input{appendices/files}

\end{document}

%% file: abstract.tex
\begin{abstract}
Machine Learning (ML) has shown significant potential in various applications; however, its adoption in privacy-critical domains has been limited due to concerns about data privacy. A promising solution to this issue is Federated Machine Learning (FedML), a model-to-data approach that prioritizes data privacy. By enabling ML algorithms to be applied directly to distributed data sources without sharing raw data, FedML offers enhanced privacy protections, making it suitable for privacy-critical environments. Despite its theoretical benefits, FedML has not seen widespread practical implementation. This study aims to explore the current state of applied FedML and identify the challenges hindering its practical adoption. Through a comprehensive systematic literature review, we assess 74 relevant papers to analyze the real-world applicability of FedML. Our analysis focuses on the characteristics and emerging trends of FedML implementations, as well as the motivational drivers and application domains. We also discuss the encountered challenges in integrating FedML into real-life settings. By shedding light on the existing landscape and potential obstacles, this research contributes to the further development and implementation of FedML in privacy-critical scenarios.\\
\end{abstract}

\begin{IEEEkeywords}
Federated Machine Learning, Collaborative Data Processing, Big Data, Systematization of Knowledge
\end{IEEEkeywords}

%% file: section/introduction.tex
\section{Introduction}\label{section:introduction}
The unprecedented growth of hyperscale computing has fuelled research in distributed architectures for training Machine Learning (ML) models at scale. Standard approaches require collecting large amounts of training data on a central server. These centralized platforms not only put the privacy of individual users at risk but also prevent cooperation between organizations due to the lack of trust in service providers \cite{schomakers2020a}. At the same time, the realization that more training data is vital for improving the performance of predictive algorithms has created economic incentives to prioritize accumulating more personal and sensitive data.
To address such increasing privacy concerns, McMahan et al. \cite{mcmahan2016a} introduced Federated Machine Learning (FedML), a novel ML paradigm that allows the training of a joint ML model on decentralized data without the need for direct data sharing. The model-to-data approach not merely increases privacy by design. Since the ML models are updated locally, the training and usage of these models can be performed without the need to communicate with a server. Additional to this offline usage, only gradients are shared, which potentially reduces the communication load compared to centralized approaches, where whole datasets are exchanged.
Despite its advantages and established theoretical framework, FedML is only sparsely adopted in real-world scenarios. This research aims to investigate the missing operationalization of FedML and disclose the challenges inhibiting a broad practical adoption. A growing literature corpus demonstrates the applicability of FedML in real-world scenarios, which provides detailed insights into the current state of applied FedML. By reviewing and systemizing this literature corpus on the described motivations, application domains, and experienced challenges, we aim to assess the current state of applied FedML. Finally, we intend to disclose the challenges inhibiting the broad practical adoption of FedML. Summarized, this systematic literature review aims to answer the following research questions (RQs):

\begin{itemize}
\item \textbf{RQ1}: What are the characteristics and trends of applied FedML?
\item \textbf{RQ2}: What are the motivational drivers to implement FedML in real-life settings and their corresponding application domain?
\item \textbf{RQ3}: What are the current challenges inhibiting the practical adoption of FedML?
\end{itemize}

%% file: section/preliminaries.tex
\section{Preliminaries}\label{section:relatedWork}
\begin{figure}[htbp]
\centerline{\includegraphics[width=0.9\columnwidth]{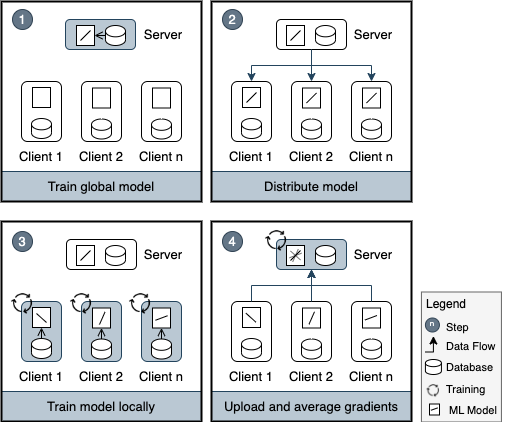}}
\caption{One iteration of the Federated Machine Learning process.}
\label{fig:FedMLProcess}
\end{figure}

FedML enables learning a joint predictive model by $K$ parties with disjoint datasets $D_{(i=1)}^K$. Traditional ML approaches require the parties to assemble their datasets in a central location, exposing data to each other and the central server. Through FedML, data owners can train a model $M_{FED}$  iteratively without the need to disclose their data. As illustrated in Figure \ref{fig:FedMLProcess}, the FedML process can be divided into four distinct steps:
\begin{enumerate}
    \item The server chooses an initial global ML model M suitable for the use case and underlying data structure. The server can pre-train the model prior to step 2.
    \item The server distributes to global model M amongst all clients.
    \item Each client $k$ computes the average gradient $g_k=\Delta F_k (w_t)$ on its local data $n_k$ at the current model $w_t$ and iterates multiple times over the update.
    \item Each client submits the gradients to the central server, which aggregates the updates from the parties.
\end{enumerate}

While classic federated learning operates on a client-server architecture, alternatives that do not rely on a central orchestrating server are also possible. For instance, parties can exchange model updates by establishing a peer-to-peer network, increasing the security of the process at the expense of consuming more bandwidth and resources for encryption \cite{roy2019a}. Moreover, the distribution of features and samples across datasets may not be homogeneous. Horizontal Federated Learning refers to the set up in which all datasets contain different samples that share the same feature space. If instead, the same samples are present in all datasets, but feature spaces are disjoint, the setup is known as Vertical Federated Learning.

%% file: section/research_process.tex
\section{Research Approach}
To provide an overview of the current state of applied FedML, we aimed to collect, structure, and summarize the existing literature through a systematic literature review (SLR). We chose an SLR approach since the current literature supports the reflection of praxis on the applicability of FedML in practical settings and since it provides a rigorous and auditable methodology that can be reviewed and replicated \cite{kitchenham2004a}. More specifically, we followed a search strategy as described by Zhang et al. \cite{zhang2011a} to identify the most relevant literature. Accordingly, our search is divided into a base literature search, a main search, and a backward search.\\

\textbf{Base Literature Search.} Our base literature consists of 7 papers. Using the base literature, we compiled a list of keywords and synonyms intended to identify literature on practical applications of FedML. As seen in Table \ref{tab:search_strings}, the identified keywords were organized into three substrings (S1-S3) which were subsequently combined into one final search string (S).

\begin{table}[!hb]
  \centering
  \begin{tabular}{p{0.5cm} p{7.3cm}}
    \toprule
    String & Query\\
    \midrule
    \textbf{S1} & Practical OR Practice OR Applied OR Real-World OR Real-Life\\
    \textbf{S2} & Federated OR Collaborative OR Decentralized OR Distributed OR Cross-Company OR Privacy-Preserving OR Multi-Party OR Secure OR Interoperable OR Cross-Industry OR Multi-Institutional\\
    \textbf{S3} &  ((Deep OR Machine) AND Learning) OR (ML OR AI OR DL) \\
    \bottomrule
    \textbf{S} & S1 AND S2 AND S3\\
  \end{tabular}\\
\caption{Compiled Search Strings for Database Search}
\label{tab:search_strings}
\end{table}

\textbf{Main Search.} To get a holistic overview of all relevant publications, we selected the most relevant seven electronic databases with a focus on computer science or software engineering \cite{chen2010a}. The list of searched databases comprises IEEE Xplore, ACM Digital Library, ISI Web of Science, Science Direct, Springer Link, Wiley InterScience, and SCOPUS. We defined the time frame from shortly after the first emergence of FedML in 2016 \cite{mcmahan2016a} to the starting date of our SLR execution in November 2021. We only included relevant literature by filtering for peer-reviewed, completed research papers in English or German with full-text access. Also, only publications within the field of computer science and technology (coarse focus), as well as work on applied FedML (narrow focus) were included. This also implies that studies that do not satisfy all inclusion criteria were excluded from the corpus. With the defined search strings and time period, we collected 3467 distinct publications (5273 including duplicates), which were independently filtered on the inclusion/exclusion criteria by two researchers. Successively, both researchers filtered solely by title (374 out of 3467), abstract (124 out of 374), and full text (70 out of 124). After every step, both researchers had an informed discussion to resolve conflicts, leading to a common list of publications for the next filtering phase.\\

\textbf{Backward Search.} We scanned the references of the resulting 70 publications. The referenced publications were filtered by title, abstract, and body according to the inclusion and exclusion criteria. The study was added if it fits all criteria and was new to the literature corpus. Through this, we added another four studies to the corpus resulting in a total selection of 74 publications. The overall search included every study from the base literature, and only four studies were added after the backward search. This indicates that the choice of search terms was comprehensive and that the search yielded a holistic overview of current literature around applied FedML.\\

\textbf{Screening Process.} To systematize the knowledge contained in the compiled literature corpus, three researchers independently extracted key information in a structured manner. To reduce the degree of bias, one researcher was not included in the SLR process and had no prior knowledge of the identified literature. Every researcher filled out a predefined data extraction sheet, which was targeted to assess the current state of applied FedML. The screening was particularly focused on the application domain, motivation, challenges, and limitations, but also on used technologies, the publication year, originating country, publication channel, research type, and research outcome. Afterward, all researchers had an informed discussion about their findings to resolve potential conflicts. The following chapters are based on the knowledge extracted from the publications identified through the SLR. An overview of the research process is shown in Figure \ref{fig:SLRProcess} of the Appendix.

%% file: section/overview_results.tex
\section{Results}
The following subsections present the findings of the data extraction as well as the systematization process and are structured according to the formulated RQs.

\subsection{Characteristics and Trends}
\begin{figure}[htbp]
\centerline{\includegraphics[width=0.9\columnwidth]{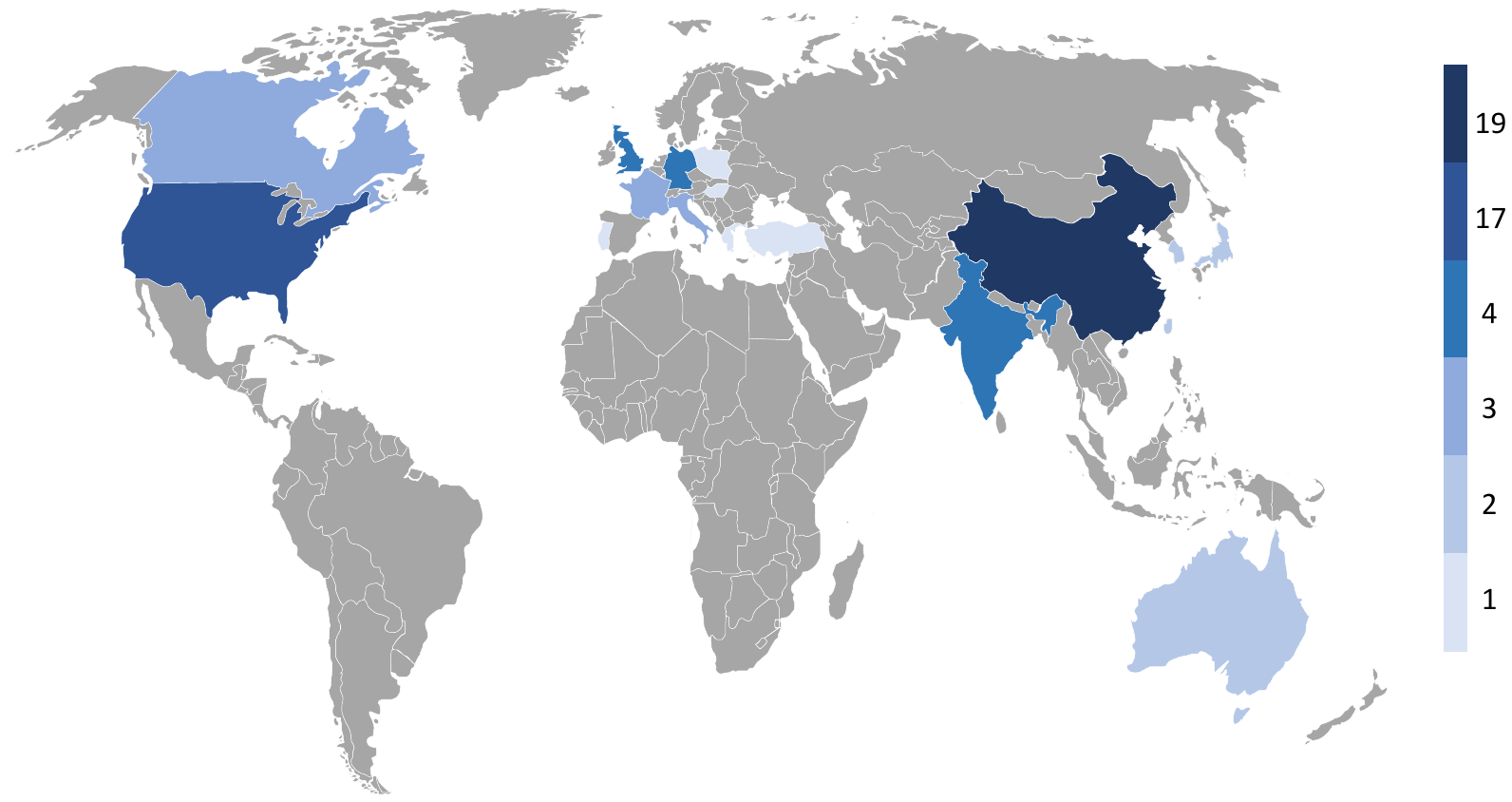}}
\caption{Distribution of Papers by Affiliated Country.}
\label{fig:global_distribution}
\end{figure}

This section aims towards answering RQ1 on the characteristics and trends of applied FedML. We started by examining the distribution of the published papers on applied FedML, research channels, affiliated countries, and research types. Then, we analyzed the technological aspects of the solution proposal to investigate potential trends.\\
While the first emergence of FedML was in 2016, the first studies on applied FedML in our literature corpus were published in 2018. The number of publications doubled from 2018 to 2019, nearly quadrupled from 2019 to 2020, and doubled again from 2020 to 2021. Due to the novelty of the field and the significant rise in research interest within the last years, the majority of our identified studies originated in 2020 and 2021. The majority of the identified studies were published at a conference (64,8\%), followed by journals (28,4\%) and workshops (6,2\%). Overall, the literature corpus comprised a diverse set of publications from 20 affiliated countries. As illustrated in Figure \ref{fig:global_distribution}, China, with a majority of 25,7\% in the overall corpus ranks first, followed by the United States with 23\%. After a large gap, Germany and India come in third with respectively 5,4\%, followed by Italy, Sweden, and France in fourth with 4\% each. Regarding the research type, most publications proposed a technological solution by demonstrating their advantages and applicability (71,6\%), followed by evaluation research approaches (16,2\%) and case studies (12,2\%). While the total amount of literature publications on applied FedML with 74 publications is relatively low, there seems to be a significant rise in interest within the last few years. Most studies seem to originate from China, focus on solution proposals, and are published at conferences.

We observed that practitioners tend to use standard neural network architectures as predictive models. Especially deep neural networks, convolutional neural networks, and autoencoders were popular choices among the authors. More sophisticated algorithms (e.g., Long-Short-Term Memory networks), or other ML models (e.g., Decision Trees) were scarcely seen. A considerable proportion (33,8\%) of the literature implemented privacy-enhancing technologies (PETs) on top of FedML to enhance privacy. Differential Privacy seems to be the most widely used PET, followed by Secure Multi-Party Computation and Homomorphic Encryption. Summarized, there seems to be a trend toward neural network architectures and the additional usage of DP to further alleviate privacy concerns. Otherwise, there seems to be no indication of potential trends.

\subsection{Motivational Drivers and Application Domains}
This section aims to answer RQ2 on the motivational drivers to implement FedML in real-life settings and their corresponding application domains. We start by analyzing the motivational drivers behind the usage of FedML in practice and then subsequently systemize the application domains.\\

\textbf{Motivational Drivers.} Although most publications aim to protect the company’s intellectual property (IP) or individuals' privacy at some point, the underlying motivation to enhance data privacy differs in each case. Sometimes, the motivation to implement privacy measures is actually driven by the motivation to overcome these data silos. We call these underlying reasons to protect data privacy motivational drivers. We identified six clusters of main motivational drivers: Privacy Protection, overcoming Data Silos, enhancing Communication Efficiency, IP Protection, Legal, and Financial. The appended Table \ref{tab:motivational_drivers} presents an cluster overview with representative papers.

\begin{figure}[htbp]
\centerline{\includegraphics[width=0.9\columnwidth]{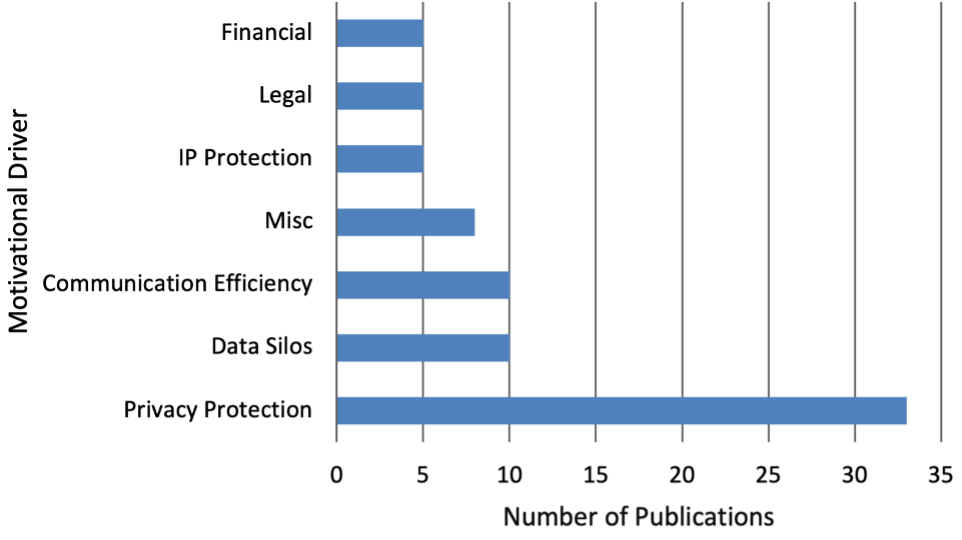}}
\caption{Distribution of Papers by Motivational Drivers.}
\label{fig:motivational_drivers}
\end{figure}

Figure \ref{fig:motivational_drivers} visualizes the distribution of the identified publication within our literature corpus by motivational drivers. The biggest part (44.6\%) of the identified literature stated to use FedML to protect the privacy of individuals without any other secondary goal. Overcoming data silos and enhancing communication efficiency were the second largest motivational drivers found in 13.5\% of the literature and followed by IP protection (6.75\%), legal reasons (6.75\%), and lastly financial concerns (4.05\%). Therefore, FedML seems to be relevant for use cases when overcoming data silos with sensitive data can alleviate data scarcity or limited communication bandwidth.

\begin{figure}[htbp]
\centerline{\includegraphics[width=0.9\columnwidth]{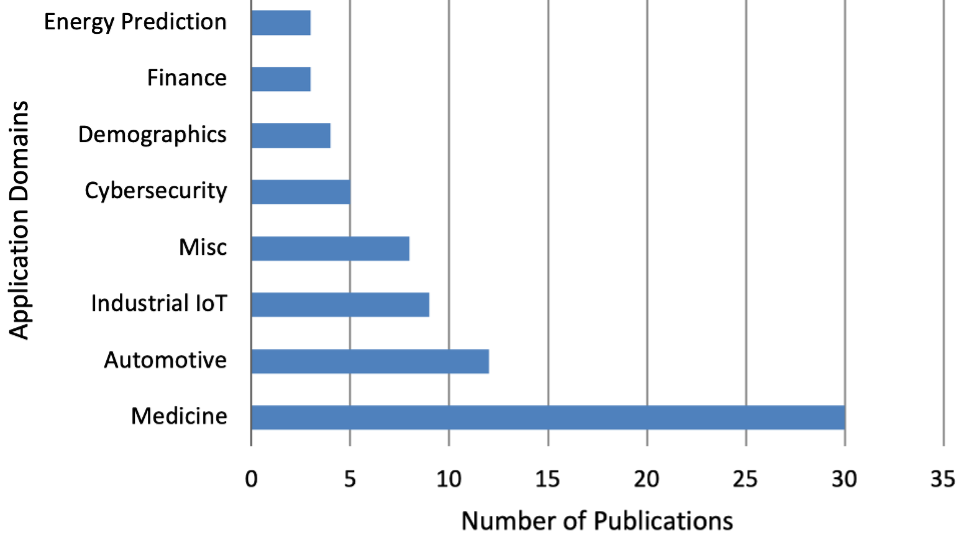}}
\caption{Distribution of Papers by Application Domains.}
\label{fig:application_domains}
\end{figure}

\textbf{Application Domains.} Similar to the previous section, we identified seven main application domains to understand in which areas FedML is especially relevant and its use could be particularly beneficial. We identified seven distinct application domains: Medicine, Finance, Automotive, Industrial Internet of Things (IoT), Energy Prediction, Demographics, and Cybersecurity. Additionally, we extracted the use cases from the literature and mapped the use cases to the corresponding domain. The results can be seen in Table \ref{tab:application_domains}. Figure \ref{fig:application_domains} displays the distribution of papers by application domain.

\begin{table}[htb]
\begin{tabular}{|c|c|c|}
\hline
\textbf{\begin{tabular}[c]{@{}c@{}}Application\\ Domain\end{tabular}}        & \textbf{Use Cases}                                                             & \textbf{\begin{tabular}[c]{@{}c@{}}Representative\\ Papers\end{tabular}}                                                     \\ \hline
\multirow{5}{*}{Medicine}                                                    & \begin{tabular}[c]{@{}c@{}}MRI or X-ray\\ Image Classification\end{tabular}    & \begin{tabular}[c]{@{}c@{}}Guo et al. \cite{guo2021a}\\ Kaissis et al. \cite{kaissis2021a}\\ Roth et al. \cite{roth2020a}\end{tabular}                          \\ \cline{2-3} 
                                                                             & \begin{tabular}[c]{@{}c@{}}Research on\\ Drug Discovery\end{tabular}           & Xiong et al. \cite{xiong2022a}                                                                                                           \\ \cline{2-3} 
                                                                             & \begin{tabular}[c]{@{}c@{}}Cancer or Tumour\\ Detection\end{tabular}           & \begin{tabular}[c]{@{}c@{}}Codella et al. \cite{codella2019a}\\ Sheller et al. \cite{sheller2018a}\\ Shen et al. \cite{shen2021a}\end{tabular}                      \\ \cline{2-3} 
                                                                             & \begin{tabular}[c]{@{}c@{}}Cardiac Health\\ Monitoring\end{tabular}            & \begin{tabular}[c]{@{}c@{}}Can \& Ersoy \cite{can2021a}\\ Ogbuabor et al. \cite{ogbuabor2021a}\end{tabular}                                           \\ \cline{2-3} 
                                                                             & Health Information                                                             & Han et al. \cite{han2021a}                                                                                                             \\ \hline
\multirow{2}{*}{Finance}                                                     & \begin{tabular}[c]{@{}c@{}}Loan Default\\ Detection\end{tabular}               & Shingi \cite{shingi2020a}                                                                                                                 \\ \cline{2-3} 
                                                                             & Fraud Detection                                                                & \begin{tabular}[c]{@{}c@{}}Byrd \&\\ Polychroniadou \cite{byrd2020a}\\ Yang et al. \cite{yang2019a}\end{tabular}                                   \\ \hline
\multirow{3}{*}{Automotive}                                                  & Autonomous Driving                                                             & \begin{tabular}[c]{@{}c@{}}Huang et al. \cite{huang2021a}\\ Jallepalli et al. \cite{jallepalli2021a}\\ Ye et al. \cite{ye2020a}\\ Zhang et al. \cite{zhang2021a}\end{tabular} \\ \cline{2-3} 
                                                                             & Traffic Prediction                                                             & \begin{tabular}[c]{@{}c@{}}Liu, Yu et al. \cite{liu2020b}\\ Q. Zhang et al. \cite{zhang2021c}\end{tabular}                                         \\ \cline{2-3} 
                                                                             & \begin{tabular}[c]{@{}c@{}}Vehicle-to-Vehicle\\ Communication\end{tabular}     & Barbieri et al. \cite{barbieri2021a}                                                                                                        \\ \hline
\multirow{3}{*}{Industrial IoT}                                              & Smart Manufacturing                                                            & \begin{tabular}[c]{@{}c@{}}Kanagavelu et al. \cite{kanagavelu2021a}\\ Lu et al. \cite{lu2020a}\end{tabular}                                            \\ \cline{2-3} 
                                                                             & Network Flow                                                                   & Kelli et al. \cite{kelli2021a}                                                                                                           \\ \cline{2-3} 
                                                                             & \begin{tabular}[c]{@{}c@{}}Unmanned Aerial\\ Vehicles (UAV)\end{tabular}       & \begin{tabular}[c]{@{}c@{}}Reus-Muns \&\\ Chowdhury \cite{reus-muns2021a}\end{tabular}                                                       \\ \hline
\multirow{2}{*}{\begin{tabular}[c]{@{}c@{}}Energy\\ Prediction\end{tabular}} & \begin{tabular}[c]{@{}c@{}}Electric Load\\ Forecasting\end{tabular}            & Taik \& Cherkaoui \cite{taik2020a}                                                                                                      \\ \cline{2-3} 
                                                                             & \begin{tabular}[c]{@{}c@{}}Energy Prediction of\\ Smart Buildings\end{tabular} & \begin{tabular}[c]{@{}c@{}}Dasari et al. \cite{dasari2021a}\\ Hudson et al. \cite{hudson2021a}\end{tabular}                                            \\ \hline
\multirow{2}{*}{Demographics}                                                & \begin{tabular}[c]{@{}c@{}}Services for\\ Communities\end{tabular}             & Hu et al. \cite{hu2019a}                                                                                                              \\ \cline{2-3} 
                                                                             & City Development                                                               & \begin{tabular}[c]{@{}c@{}}Y. Liu, Huang et al. \cite{liu2020a}\\ Mashhadi et al. \cite{mashhadi2021a}\end{tabular}                                   \\ \hline
\multirow{5}{*}{Cybersecurity}                                               & False Data Injection                                                           & Zhao et al. \cite{zhao2022a}                                                                                                            \\ \cline{2-3} 
                                                                             & Intrusion Detection                                                            & Sun et al. \cite{sun2021a}                                                                                                             \\ \cline{2-3} 
                                                                             & Fault Detection                                                                & X. Wang et al. \cite{wang2021a}                                                                                                         \\ \cline{2-3} 
                                                                             & Fraud Detection                                                                & Marulli et al. \cite{marulli2021a}                                                                                                         \\ \cline{2-3} 
                                                                             & \begin{tabular}[c]{@{}c@{}}Network Traffic\\ Security\end{tabular}             & He et al. \cite{he2021a}                                                                                                              \\ \hline
Misc.                                                                        & Diverse                                                                        & \begin{tabular}[c]{@{}c@{}}Patel et al. \cite{patel2021a}\\ Ruan et al. \cite{ruan2021a}\end{tabular}                                               \\ \hline
\end{tabular}
\caption{List of Application Domains, their Uses Cases, and Representative Papers.}
\label{tab:application_domains}
\end{table}

With 40.54\%, most publications on applied FedML were in the field of medicine and healthcare, followed by the automotive sector (16.22\%) and Industrial IoT (12.17\%). Less work has been done in the field of cybersecurity (6.75\%), demographics (5.4\%), finance, and energy prediction (4.05\%). Hence, FedML seems to be especially relevant for application domains either with privacy-sensitive user data, IP-sensitive data, or settings with limited computational power and communication bandwidth.

\begin{figure*}[htbp]
\centerline{\includegraphics[width=\columnwidth*2]{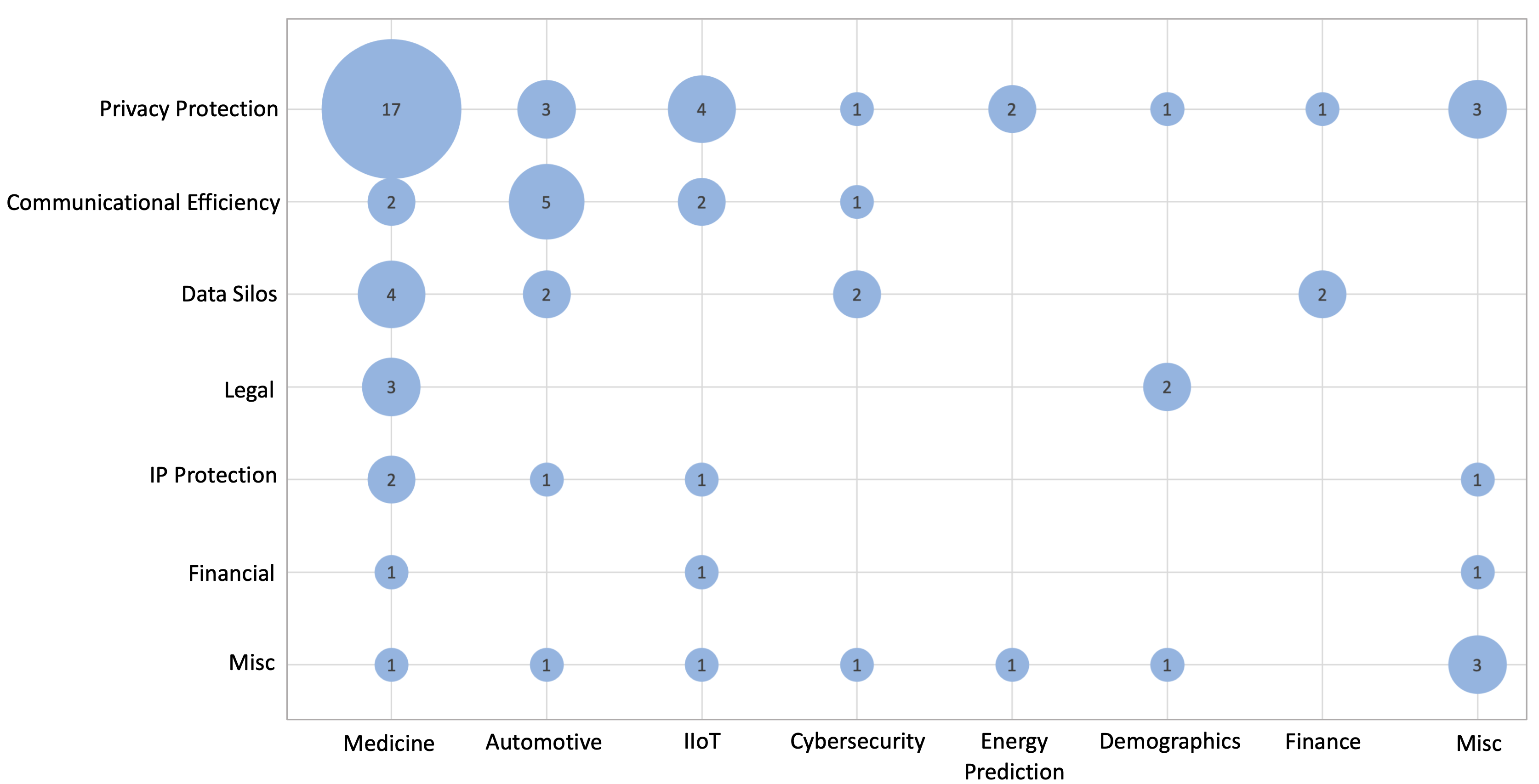}}
\caption{Mapping of Motivational Drivers and Application Domain from Literature Corpus.}
\label{fig:mapping}
\end{figure*}

Finally, we investigated which motivational drivers for each application domain to get a better understanding of why different domains apply FedML in practice. As illustrated in Figure \ref{fig:mapping}, the mapping landscape is diverse. The most distinct intersection is between privacy protection and the medicine domain. Notably, the automotive domain is mainly motivated to enhance communication efficiency by leveraging the model-to-data approach of FedML. The Industrial IoT (IIoT) domain mostly uses FedML to protect privacy and improve communicational efficiency. Otherwise, no distinct intersection can be observed. Privacy Protection is a general motivational driver throughout each application domain. Communicational efficiency seems to be an irrelevant motivation for cybersecurity, energy prediction, demographic, and finance use cases. The financial motivational driver was only reported in the medicine and IIot domain. Similarly, only the medicine and demographics domain reported being motivated by legal compliance, and IP protection motivation was only described by studies from the medicine, automotive, and IIoT domain.

This observation can be attributed to the relatively small literature corpus and might change if more studies emerge on applied FedML. In summary, there seems to be no current distinct intersection except for the usage of FedML in the medicine domain for privacy protection and in the automotive sector to increase communication efficiency.

\subsection{Open Challenges}
This section aims to answer RQ3 on the current challenges inhibiting the practical adoption of FedML. The novelty and hype around FedML sparked huge research efforts yielding inspiring work for a large range of possible challenges of FedML systems \cite{khan2021,SoK2021}. However, a narrower focus on a subset of all possible challenges is needed if we aim to bridge the gap from theoretical FedML to practical application. Applying technologies in production settings unavoidably introduces restrictions like limited bandwidth or faulty hardware. Theoretically, these limitations could be artificially implemented and considered but might still differ heavily in real-life settings. Against this backdrop, we intend to put a spotlight on these specific challenges. Based on the identified literature corpus, we systemized the described challenges in the practical adoption of FedML.\\

\textbf{Challenge 1: System Heterogeneity.} Most FedML systems work well for independent and identically distributed (IID) data. However, having IID data across a multitude of devices is hard to achieve, especially when several different institutions are involved. This systems heterogeneity posed the main challenge in most of the identified publications \cite{fan2021a,liu2020b,mashhadi2021a,zhuRobustFederatedLearning2020}. The problem of training data heterogeneity reaches from data imbalance in statistical distribution and size \cite{marulli2021a,roth2020a}, inconsistent label formats or missing labels \cite{jallepalli2021a} to class imbalances \cite{roth2020a}. However, some authors encountered further heterogeneity issues, such as heterogeneous hardware settings \cite{ek2021a} and different network environments \cite{zhang2021a}.\\

\textbf{Challenge 2: Scalability.} Most publications in our literature corpus applied FedML for a setting with a relatively small client size (approx. 3-5 clients). As reported, FedML works well for smaller settings. The challenge occurs when the authors increase the number of participating parties. With a larger set of clients, the overall size of the dataset and the needed communication bandwidth also increases. Naturally, the systems have to accommodate larger communicational overhead and data load \cite{dasari2021a,hudson2021a,kanagavelu2021a,zhang2021b}. Some authors also described reduced accuracy when the number of clients was increased \cite{ahmed2020a}. Additionally, a bigger set of clients could also introduce a higher risk of system heterogeneity issues as described in Challenge 1.\\

\textbf{Challenge 3: Privacy.} Privacy is a core challenge of FedML systems. Information can be learned by inspecting the global shared model parameters, and sensitive data can be learned by gaining access to the updates of an individual institution \cite{codella2019a,liu2021d,yang2019a}. Some authors counteracted the problem by adding additional PETs. However, applying these PETs introduces further challenges, like the privacy-utility trade-off \cite{guo2018a,lu2020a}, as well as additional computation and communication overhead \cite{kaissis2021a}. These communication burdens could be partially alleviated by research on hardware restrictions or a more efficient communication protocol.\\

\textbf{Challenge 4: Hardware Restrictions.} Hardware constraints are a challenge that gains importance, especially in real-world deployments. Insufficient computational power or restricted bandwidth could be the potential bottleneck for training the local ML models \cite{can2021a,liu2021d,liu2020b,taik2020a}. This local training also implies that edge devices need to have enough computational power to actually run these ML models. If the computation power of these edge devices is not sufficient, the ML models need to be smaller, which might reduce their predictive performance \cite{majeed2020a,zhang2021b}. Some use cases require sophisticated ML models and are therefore potentially restricted by limited computational power. In these cases, the bottleneck for a successful application is the hardware.\\

\textbf{Challenge 5: Communication Protocol.} The decentralized nature of FedML requires setting up a more sophisticated communication protocol compared to traditional centralized ML approaches. The required organization and hardware to set up the communication protocol yields a multitude of challenges \cite{barbieri2021a,cetinkaya2021a,ek2021a,kanagavelu2021a}. Especially when scaling to a larger client size, the communication costs are exponential to the number of peers, which consequently also increases execution time and overhead while decreasing efficiency. This communication protocol poses a bottleneck in some studies. Additionally, one study describes a limit to the update frequency \cite{fan2021a}. From an organizational perspective, some authors encountered problems synchronizing the aggregation of client model updates. Hence, there is a trade-off between communication costs and network load.

%% file: section/conclusion.tex
\section{Discussion}
Through a systematic literature review, we identified 74 publications on the practical adoption of FedML. We analyzed the literature and assessed the current state of applied FedML. We systemized the motivational drivers, application domains and encountered challenges. 
Our study yielded several insights. We observed that the interest in applied FedML has risen significantly over the last few years since its emergence in 2016. The analysis of affiliated countries suggests that China and the United States are the most active countries involved in shaping the research on applied FedML. Especially the health domain seems to benefit from the privacy advantages of FedML, but also the automotive industry indicates growing FedML investment due to increased communication efficiency and enhanced data privacy. Practitioners tend to use standard neural network architectures, such as DNNs or CNNs as the preferred predictive models and in approximately 33\% of the use cases additionally implement PETs to further increase data privacy. The report suggests that the application of FedML in real-life settings is currently mainly inhibited by the heterogeneity of the underlying systems, followed by scalability issues, privacy concerns, hardware restrictions, and a bottleneck in the communication protocol.
The overall search results suggest that the choice of search terms was comprehensive and that the search yielded a holistic overview of the current literature. However, we might have missed certain search terms and therefore also relevant studies, which might have altered the results of our systematization. Additionally, the screening and systematization process might be influenced by subjective bias. We tried to counteract this through regular discussions, predefined selection criteria, and introducing an additional researcher who was not part of the search process. Furthermore, the list of identified challenges concentrates on the technical aspects of applied FedML due to the technical nature of the identified research papers. Our study solely represents the current state of the literature with a relatively small sample size of 74 publications due to the novelty of FedML. We expect our results to change with every novel study. Our study contributed to the understanding of the current state, characteristics, and challenges of applied FedML. However, future use cases will make the impact of FedML in real-world settings more apparent.
Real-life settings introduce another dimension of challenges since human decisions are crucial for the practical adoption of a novel technology. Especially the decentralized nature of FedML requires collaboration. Setting up and coordinating collaborative settings poses another potential reason for failure and currently unknown challenges that might need to be tackled. We encourage researchers not only to focus on the described technical problems but also to investigate the social and socio-technical challenges of collaborative ML.

%% file: bibliography.tex

%% file: appendices/files.tex
\subsection{Research Process}

\begin{figure}[htbp]
\centerline{\includegraphics[angle=90,width=0.95\columnwidth]{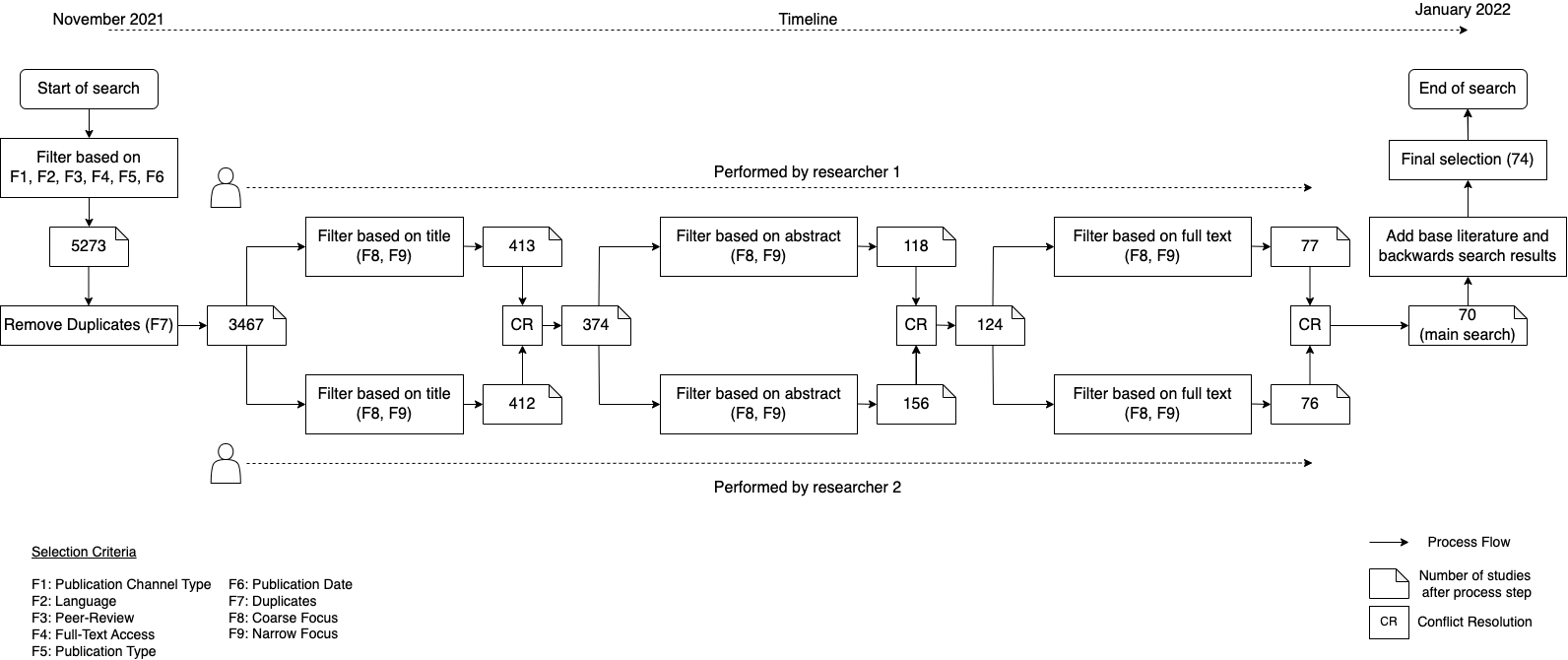}}
\caption{Overview of the Research Process.}
\label{fig:SLRProcess}
\end{figure}
\newpage

\subsection{Overview of Motivational Drivers.}

\begin{table}[H]
\centering
\rotatebox{90}{
\begin{tabular}{|c|l|c|}
\hline
\textbf{Motivational Driver}                                       & \multicolumn{1}{c|}{\textbf{Description}}                                                                                                                                                                                                                                           & \textbf{Representative Papers}                                                                          \\ \hline
Financial                                                          & \begin{tabular}[c]{@{}l@{}}Financial comprises literature with the motivational driver of cost-related concerns,\\ e.g., for evading potential financial penalties.\end{tabular}                                                                                                    & \begin{tabular}[c]{@{}c@{}}Cetinkaya et al. \cite{cetinkaya2021a}\\ Lu et al. \cite{lu2020a}\end{tabular}                        \\ \hline
Legal                                                              & \begin{tabular}[c]{@{}l@{}}This cluster includes literature where legal  regulation (e.g., GDPR) obliged practitioners\\ to set up privacy-enhancing methods.\end{tabular}                                                                                                          & \begin{tabular}[c]{@{}c@{}}Choudhury et al. \cite{choudhury2020a}\\ Hu et al. \cite{hu2019a}\\ Liu et al. \cite{liu2021d}\end{tabular}     \\ \hline
\begin{tabular}[c]{@{}c@{}}Privacy\\ Protection\end{tabular}       & \begin{tabular}[c]{@{}l@{}}This cluster incorporates literature that applies FedML to protect individuals' privacy.\\ The motivational driver is the social factor of protecting the well-being of individuals\\ and not protecting privacy to comply with e.g., GDPR.\end{tabular} & \begin{tabular}[c]{@{}c@{}}Fan et al. \cite{fan2021a}\\ Fang et al. \cite{fang2021a}\\ Hegedűs et al. \cite{heged2020a}\end{tabular}     \\ \hline
IP Protection                                                      & This cluster focuses on protecting IP of e.g., companies instead of individuals' privacy.                                                                                                                                                                                           & \begin{tabular}[c]{@{}c@{}}Kumar et al. \cite{kumar2021a}\\ Majeed et al. \cite{majeed2020a}\\ Mercier et al. \cite{mercier2021a}\end{tabular} \\ \hline
Data Silos                                                         & \begin{tabular}[c]{@{}l@{}}To alleviate data scarcity, organizations can try to overcome data silos. This cluster\\ comprises literature where FedML was leveraged to overcome data silos.\end{tabular}                                                                             & \begin{tabular}[c]{@{}c@{}}Shingi \cite{shingi2020a}\\ Wang et al. \cite{wang2020a}\\ Zhao et al. \cite{zhao2022a}\end{tabular}            \\ \hline
\begin{tabular}[c]{@{}c@{}}Communication\\ Efficiency\end{tabular} & \begin{tabular}[c]{@{}l@{}}When a central cloud must be queried, the cost of data transmission is high, and scalability\\ issues occur. This cluster incorporates studies that used FedML to optimize communication\\ and bandwidth efficiency\end{tabular}                         & \begin{tabular}[c]{@{}c@{}}Ek et al. \cite{ek2021a}\\ Sacco et al. \cite{sacco2020a}\end{tabular}       \\ \hline
Miscellaneous                                                      & Work, where the motivational driver cannot be clearly assigned to other clusters.                                                                                                                                                                                                   & Ahmed et al. \cite{ahmed2020a}                                                                                      \\ \hline
\end{tabular}
}
\caption{Overview of Motivational Drivers, their Description, and Representative Papers.}
\label{tab:motivational_drivers}
\end{table}